\documentclass[letterpaper, 10 pt, conference]{ieeeconf}  

\IEEEoverridecommandlockouts                              

\usepackage{amsmath} 
\usepackage{amssymb}  

\usepackage{booktabs}
\usepackage{graphicx}
\usepackage{caption}
\usepackage{cuted}
\usepackage{algorithm}
\usepackage{algpseudocode}
\usepackage{hyperref}
\usepackage{comment}

\title{\LARGE \bf
FlyMirage: A Fully Automated Generation Pipeline for Diverse and Scalable UAV Flight Data via Generative World Model
}
\author{
Jinhan Li$^{\#,1,2}$,
Xijie Huang$^{\#,1,2}$,
Weiqi Gai$^{2}$,
Zhaoqi Wang$^{2}$,
Qiyi He$^{2}$,
Yijin Wang$^{1,2}$,\\
Mo Zhu$^{1,2}$,
Zhiyang Liu$^{2}$,
Fei Gao$^{1,2}$,
Xin Zhou$^{2}$,
Yuze Wu$^{*,1,2}$,
Ziming Ding$^{2}$%
\thanks{$^{1}$Zhejiang University , $^{2}$Differential Robotics}%
\thanks{$^{\#}$Equal contribution}%
\thanks{$^{*}$Corresponding author: wuyuze000@zju.edu.cn}%
}

\begin{document}

\makeatletter
	\let\@oldmaketitle\@maketitle
	\renewcommand{\@maketitle}{\@oldmaketitle
			\includegraphics[width=1.0\linewidth]{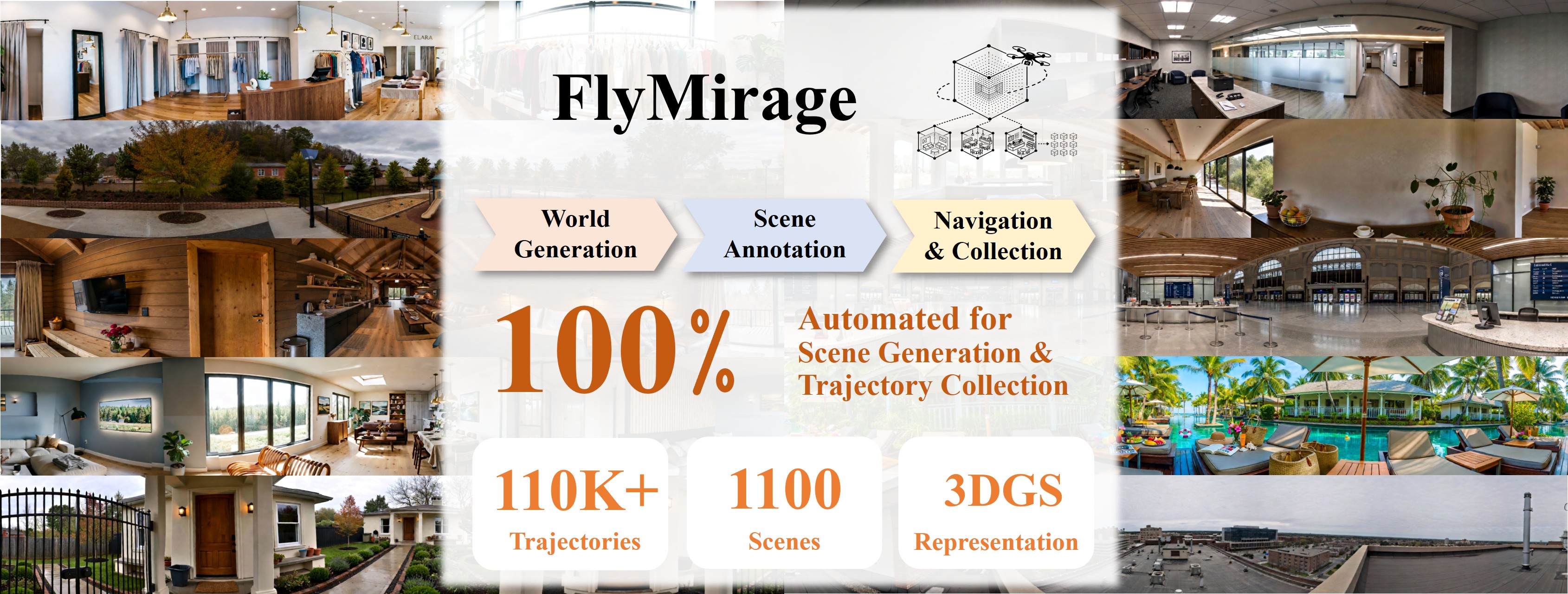}
		\captionsetup{font={small}}
		\captionof{figure}{
			\label{fig:top}
\textbf{FlyMirage}. It uses LLM-designed scene specifications, generative world model, automated scene annotation, and
UAV-feasible trajectory planning to produce scalable, diverse, and photorealistic aerial VLN data.
		}
  	}
\makeatother
\maketitle
\setcounter{figure}{1}
\thispagestyle{empty}
\pagestyle{empty}

\begin{abstract}
In the field of Vision-Language Navigation (VLN), aerial datasets remain limited in their ability to combine scale, diversity, and realism, often relying on either costly real-world scenes or visually limited simulations. To address these challenges, we introduce FlyMirage, a highly scalable and fully automated data generation pipeline for aerial VLN. Our approach leverages large language models (LLM) as an environment designer to promote scene diversity, paired with a generative world model that instantiates these designs into high-fidelity 3D Gaussian Splatting (3DGS) scenes. To substantially reduce human labor and ensure the feasibility of flight data, FlyMirage automates scene exploration and semantic information acquisition, and further integrates a dynamically feasible planner for uncrewed aerial vehicle (UAV) trajectory generation. Utilizing this toolchain, we generate a large-scale, diverse, and photorealistic aerial VLN dataset. We further fine-tune the widely used $\pi_0$ model on this dataset and demonstrate the improvement our data brings to the model's ability to navigate in completely unseen environments, both in simulation and in the real world. Project website: \url{https://lij1500.github.io/flymirage_website/}.

\end{abstract}

\section{INTRODUCTION}

Advances in embodied intelligence often depend on the availability of large-scale, high-quality training data. For example, models such as Gen-1~\cite{gen-1}, trained on approximately half a million hours of real-world data, demonstrate strong generalization capabilities. However, in the field of aerial Vision-Language Navigation (VLN), available training datasets remain limited and are primarily derived from two sources: real-world and simulated environments.

Real-world navigation data is typically collected by experienced human operators through physical drone flights, making it costly to acquire and difficult to scale. For example, UAV-Flow~\cite{wang2025uavflowcolosseorealworldbenchmark} collected 100 hours of pilot-operated flight data, but its environmental coverage remains limited to university campuses.
In contrast, simulated data enables a safer, cheaper and more scalable approach. Simulated data refers to data collected in digital environments, including real-world reconstructions~\cite{mattersim,rxr,krantz_vlnce_2020,miao2025physicallyexecutable3dgaussian}, and scenes built with virtual 3D assets. Within these simulations, prior efforts such as OpenFly~\cite{OpenFly} attempt to automate trajectory generation with A* search in 18 pre-constructed scenes, but their scalability remains bounded by the availability of high-quality predefined environments. Despite this progress, we argue that current real-world and simulated data pipelines still face three key limitations:

\begin{enumerate}
    \item \textbf{Limited visual and geometric fidelity.} Asset-based virtual environments suffer from simplified textures, materials, and lighting, increasing the visual sim-to-real gap.

    \item \textbf{Insufficient data diversity and scale.} 
   Achieving scalable diversity of high-quality scenes remains difficult for both real-world and simulation pipelines, as they often require human intervention in scene construction, scene annotation or trajectory collection.
    
    \item \textbf{Dynamics-unaware trajectory generation.}
Although prior attempts have used search-based algorithms for automated trajectory generation, they ignore robot dynamics and can produce trajectories that are unnatural or infeasible for real-world UAV deployment.
\end{enumerate}

Recent advances in generative world models offer a compelling opportunity to overcome these limitations. In particular, foundation models like Marble~\cite{worldlabs_marble_models} can generate explorable 3D worlds with coherent spatial structures from text or image prompts. This capability opens up a promising avenue for rethinking aerial VLN data generation: instead of relying on data collected in real-world environments or manually constructed virtual spaces, we can automatically generate realistic, explorable worlds and collect navigation data within them.

Building on this capability, we propose a scalable and fully automated pipeline for generating aerial VLN data (Fig. \ref{fig:top}).
To enrich scene diversity, we leverage large language models (LLM) as a navigation environment designer that generates diverse structured scene specifications, which are then instantiated by the generative world model into photorealistic, explorable 3D worlds. This design enables a dataset to cover a broad spectrum of environments, ranging from common indoor and outdoor scenes to hazardous, access-restricted, and hard-to-reconstruct spaces such as radiation-risk areas or chemical facilities.
To reduce the visual sim-to-real gap, we use Marble 1.1 Plus as the generative world model, following a text$\&$image-to-world paradigm to create high-fidelity 3D Gaussian Splatting (3DGS) \cite{kerbl3Dgaussians} scenes.
To minimize human involvement, we automate camera exploration and scene annotation. We further develop an automatic aerial navigation target-generation system based on the resulting scene annotations, and integrate a UAV-specific dynamically feasible trajectory planner for trajectory generation. We denote this fully automated pipeline as \textbf{FlyMirage}. With a single command, it can continuously generate a virtually unbounded collection of diverse scenes and corresponding navigation trajectories.

Therefore, we position FlyMirage as the next-generation tool of automated aerial VLN data collection. To validate its potential, we generate 1100 distinct example scenes from indoor to outdoor and 110K+ navigation trajectories automatically. Comparing with existing VLN datasets, FlyMirage offers superior extensibility and observation fidelity, while also producing physically executable aerial trajectories. Moreover, the monetary and time costs of data generation are substantially lower: FlyMirage requires only Marble and a consumer-grade NVIDIA GPU to render RGB images from 3DGS scenes, in contrast to the manual mesh creation required by InteriorGS \cite{InteriorGS2025} or the human-operated drone flights required in UAV-Flow \cite{wang2025uavflowcolosseorealworldbenchmark} that cost approximately \$100 per hour.

We summarize our contributions as follows:
\begin{enumerate}
    \item We introduce \textbf{FlyMirage}, a fully automated aerial VLN data generation pipeline powered by a generative world model, providing a scalable and realistic source of UAV navigation data.

    \item Using this pipeline, we construct a large-scale dataset comprising 1100 diverse scenes and over 110K aerial trajectories, offering greater scale, visual diversity, and dynamic feasibility than existing datasets.

    \item Through extensive simulation and real-world experiments, we show that training on our dataset improves existing VLA models and enhances their generalization to unseen environments.

    \item We will release our navigation dataset to facilitate future research in  embodied aerial VLN.
\end{enumerate}

\begin{figure*}[t]
  \centering
  \includegraphics[width=0.95\textwidth]{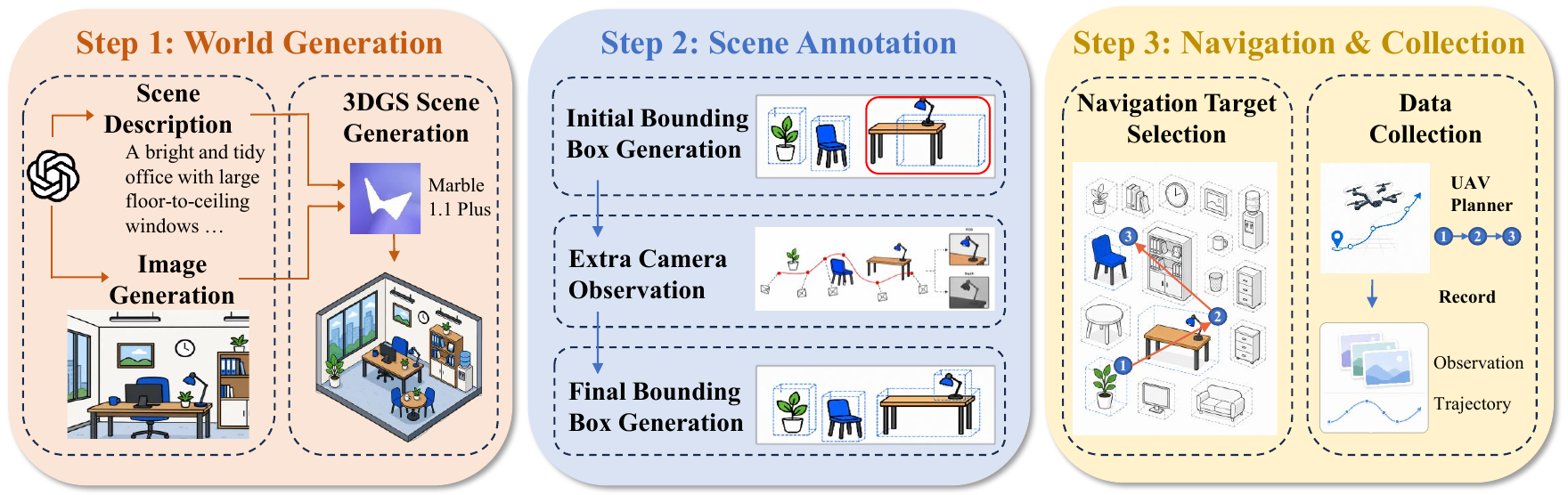}
  \caption{\textbf{Overall Pipeline}. It consists of three stages: World Generation, Scene Annotation and Navigation \& Collection.}
  \label{fig:overall}
  \vspace{-1em}
\end{figure*}

\section{Related Work}
\subsection{Labor Intensive Dataset Collection}
Vision-Language Navigation (VLN) tasks originated in ground robotics, where early works commonly relied on graph-based maps that restrict navigation to transitions between predefined nodes, enabling efficient trajectory sampling together with manual language annotation \cite{mattersim,rxr}. Such node-based formulations substantially simplify the robot's action space, but are less suitable for UAV navigation, where trajectories are expected to exploit the greater freedom of motion available in 3D space. Supporting this richer action space makes automatic trajectory collection considerably more challenging, and many aerial VLN datasets therefore rely on human piloting. For instance, UAV-Flow \cite{wang2025uavflowcolosseorealworldbenchmark} amassed 30k episodes controlled by human operators in the real world. Other efforts have crowdsourced trajectories from humans flying drones within simulated environments, including CityNAV \cite{lee2024citynav}, AerialVLN \cite{liu_2023_AerialVLN}, and AVDN \cite{fan-etal-2023-aerial}. Despite relying on manual trajectory generation, some works, such as OpenUAV \cite{wang2024realisticuavvisionlanguagenavigation}, leverage large language models to automatically generate natural language annotations for these paths.

\subsection{Automatic Dataset Collection}
Significant efforts have also been directed toward automating trajectory generation, although such approaches often still require manual 3D scene construction. To facilitate this, SAGE-3D \cite{miao2025physicallyexecutable3dgaussian} handcrafted 1,000 3DGS scenes, introduced as InteriorGS \cite{InteriorGS2025}. Based on such constructed scenes, search-based planning algorithms have become a standard approach for automatically generating trajectories in both ground \cite{krantz_vlnce_2020} and aerial robotics \cite{OpenFly}. 
While there are attempts to generate scalable 3D mesh-based scene representations~\cite{yang2025sceneweaver}, these methods primarily construct scenes by composing and arranging 3D assets. Their object-centric representations, with individually modeled and interactable objects, are particularly well suited for Vision-Language-Action (VLA) manipulation tasks. However, the visual gap between such asset-composed environments and real-world observations can limit their suitability for collecting navigation data intended for real-world deployment. While recent work \cite{dreamgen} leverages video world models to synthesize training data, it incurs a prohibitive computational cost, requiring approximately 81,000 NVIDIA L40 GPU hours to generate 240k samples.

\section{Pipeline for Dataset Creation}
As illustrated in Fig.~\ref{fig:overall}, our pipeline automatically generates scalable UAV navigation datasets by transforming batch-produced scene descriptions into UAV trajectories for training navigation policies.

\subsection{World Generation}

\begin{figure}[h]
  \centering
    \includegraphics[width=0.75\linewidth]{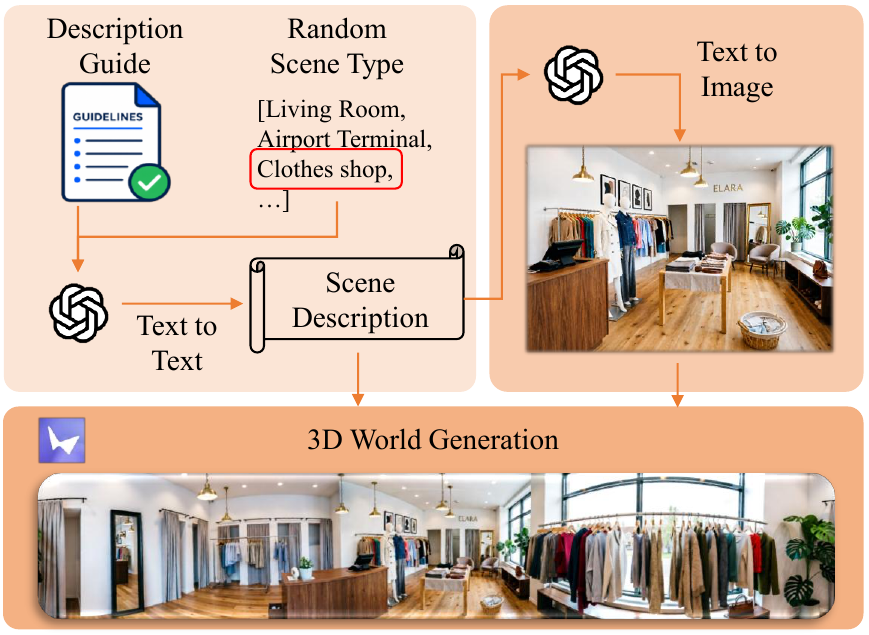}
    \caption{\textbf{World Generation}. The batch process from randomly selected scene types to complete 3DGS scenes.}
  \label{fig:stage 1}
  \vspace{-1em}
\end{figure}

We organize common types of scenes into a hierarchical taxonomy of categories and subcategories. To generate each scene description, we first randomly select a subcategory from a broader category, such as ``living room'' under ``residential spaces'' and then prompt GPT-5.4 to generate a description consistent with the selected scene type. This two-step procedure is designed to mitigate the uneven sampling behavior we observed when prompting LLMs to generate random scene descriptions directly. In particular, direct prompting often led to a biased distribution over scene types; for example, GPT-5.4 produced medical-related descriptions substantially more frequently than other common environments. This allows us to maintain a more controlled and balanced distribution of scene types while still benefiting from the model’s ability to produce diverse, detailed, and naturalistic descriptions of common real-world spaces.

Given the resulting textual description, we next generate a corresponding image using GPT Images 2.0. This image generation step provides a visually grounded representation of the scene, complementing the language-only description with concrete spatial, appearance, and layout cues. Compared with a text description alone, the generated image can capture additional visual details such as object placement, material appearance, lighting conditions, color composition, and the overall spatial arrangement of the environment. We then use both the generated image and the original textual description as inputs to Marble 1.1 Plus to generate a 3DGS-based scene, as illustrated in Fig. \ref{fig:stage 1}.

\subsection{Scene Annotation}
\begin{figure*}[htb]
  \centering
    \includegraphics[width=0.75\linewidth]{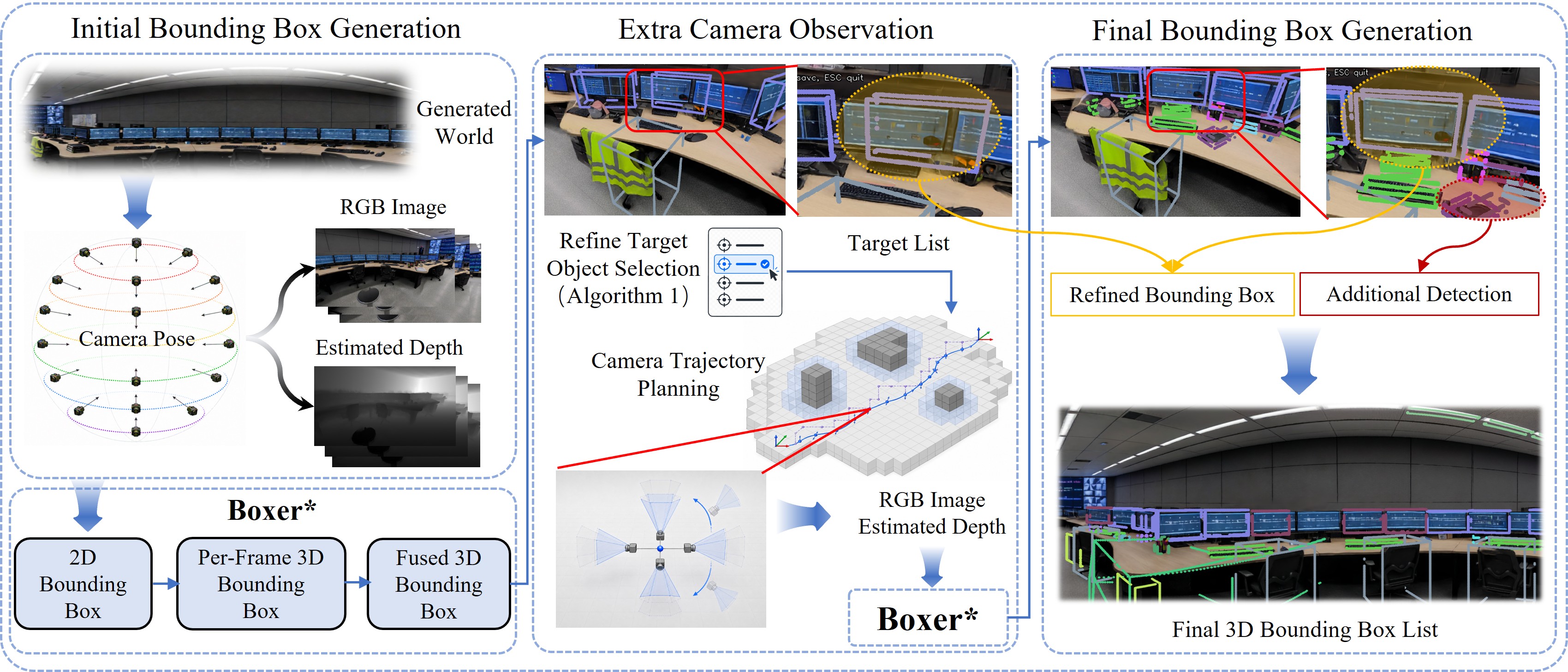}
    \caption{\textbf{Scene Annotation}. The iterative algorithm is used to generate accurate bounding boxes for objects in generated scenes. (*Boxer \cite{boxer2026} is a 3D bounding box estimation algorithm by Meta Reality Labs)}
  \label{fig:stage 2}
\end{figure*}

Marble produces only a 3DGS scene representation, without training images or object-level semantic information. To obtain object-level annotation, we use Boxer~\cite{boxer2026} to perform open-vocabulary object detection and estimate 3D bounding boxes for the detected objects. Specifically, we use GSplat~\cite{ye2025gsplat}-rendered RGB images and the corresponding estimated depth maps as inputs to Boxer. However, due to the limitation of the Boxer algorithm, the 3D bounding-box estimates become less accurate for objects that are far from the rendering viewpoint. 

Thus, we use a simple iterative method to explore the scene as shown in Fig. \ref{fig:stage 2}. Since Marble follows a ``prompt-to-panorama-to-world'' pipeline, every generated scene has a center $\mathbf{c}$, where we orbit the camera around it with radius \(r_{\mathrm{orb}}=0.1\,\mathrm{m}\). Camera viewpoints are sampled at multiple elevation angles over the viewing sphere, with a complete yaw sweep performed at each elevation, as illustrated by the camera pose visualization in Fig.~\ref{fig:stage 2}.

\begin{algorithm}[htb]
\caption{Distance-Aware Target Selection}
\label{alg:target_selection}
\begin{algorithmic}[1]
\Require Candidates $\mathcal{O}=\{\mathbf{o}_i\}_{i=1}^{N}$ with centers
$\{\mathbf{x}_i\}_{i=1}^{N}$, scene center $\mathbf{c}$, thresholds
$d_{\mathrm{th}1}, d_{\mathrm{th}2}, \theta_{\mathrm{th}}$, and target number $N_t$

\State $d_i\gets\|\mathbf{x}_i-\mathbf{c}\|_2,\ \forall i$; sort $\mathcal{O}$ by decreasing $d_i$
\State $\mathcal{T}\gets\emptyset,\quad \mathcal{A}\gets\mathcal{O}$

\For{each $\mathbf{o}_i\in\mathcal{O}$}
    \If{$\mathbf{o}_i\in\mathcal{A}$ \textbf{and} $d_i>d_{\mathrm{th}2}$}
        \State $\mathcal{T}\gets\mathcal{T}\cup\{\mathbf{o}_i\}$
        \State $\mathcal{A}\gets\mathcal{A}\setminus
        \{\mathbf{o}_j:
        \|\mathbf{x}_j-\mathbf{x}_i\|_2<d_{\mathrm{th}1}
        \ \textbf{or}\
        \Delta_{\mathrm{az}}(i,j)<\theta_{\mathrm{th}}\}$
        \If{$|\mathcal{T}|=N_t$} \State \textbf{break} \EndIf
    \EndIf
\EndFor

\If{$\mathcal{T}=\emptyset$}
    \State $\mathcal{Q}\gets
    \{\mathbf{o}_i\in\mathcal{O}: d_{\mathrm{th}1}<d_i\le d_{\mathrm{th}2}\}$
    \If{$\mathcal{Q}\neq\emptyset$}
        \State $\mathcal{T}\gets
        \{\arg\max_{\mathbf{o}_i\in\mathcal{Q}} d_i\}$
    \EndIf
\EndIf
\State \Return $\mathcal{T}$
\end{algorithmic}
\end{algorithm}

Given these camera poses, we render RGB images and estimated depth maps and feed them to Boxer. We denote the returned 3D bounding boxes as the candidate set $\mathcal{O} = \{\mathbf{o}_i\}_{i=1}^{N}$. Due to limitations of Boxer, we define two distance thresholds, \(0 < d_{\mathrm{th}1} < d_{\mathrm{th}2}\). Predicted bounding boxes for objects farther than \(d_{\mathrm{th}2}\) are often substantially inaccurate and therefore require closer camera observation, whereas bounding boxes for objects with distances between \(d_{\mathrm{th}1}\) and \(d_{\mathrm{th}2}\) may be slightly inaccurate but remain within an acceptable range. Thus, we select targets for extra camera observation as shown in Algorithm \ref{alg:target_selection}. Starting from the farthest candidate, we greedily accept candidates with \(d_i>d_{\mathrm{th}2}\). After accepting a candidate, we discard nearby objects within $d_{\mathrm{prune}}$ where $d_{\mathrm{prune}} = d_{\mathrm{th}1}$, or whose azimuth with respect to $\mathbf{c}$ differs by less than $\theta_{\mathrm{th}}$. If no candidate is selected, we fall back to closer candidates.

Given a selected target, we plan a collision-checked 3D waypoint path toward the target on an inflated occupancy grid, yielding a dense camera trajectory. A four-view camera rig is placed at fixed intervals along the trajectory, corresponding to front, left, right, and back views. Using the extra camera observations together with the initial rotational observations, we rerun a modified Boxer framework and filter out Boxer 3D detections farther than $d_{\mathrm{th}1}$ from the camera optical center in each frame. This eliminates distant artifacts and improves the fidelity of the final fused global bounding-box map.

In our implementation, to ensure both data diversity and accurate bounding-box localization, we set \(N_t =\) 5, \(\theta_\mathrm{th}=\) 35$^\circ$, \(d_{\mathrm{th1}}=\) 3.0 m, and \(d_{\mathrm{th2}}=\) 4.0 m.

\subsection{Navigation \& Collection}
\begin{figure}[htb]
  \centering
    \includegraphics[width=0.8\linewidth]{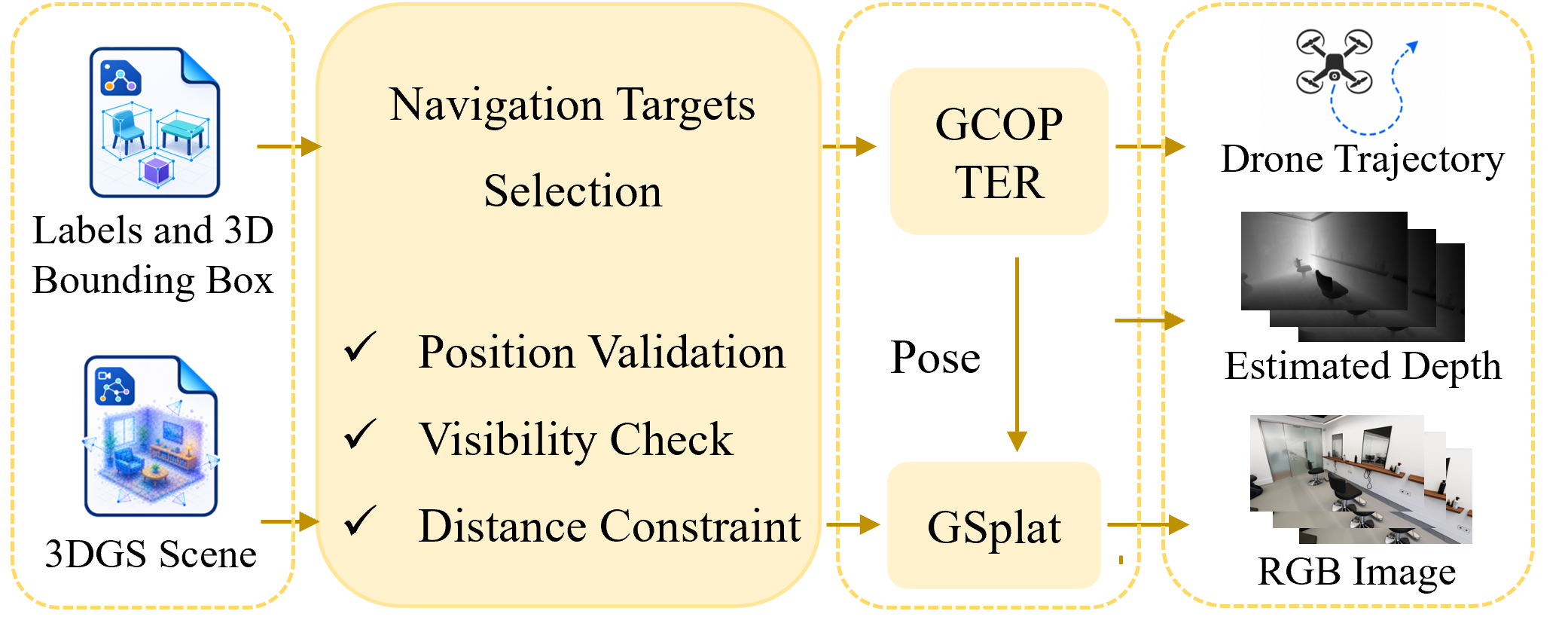}
    \caption{\textbf{Navigation \& Collection}. An automated pipeline to determine trajectory goals and deploy GCOPTER.}
  \label{fig:stage 3}
\end{figure}

As shown in Fig.~\ref{fig:stage 3}, the navigation and data collection pipeline requires only two scene-level inputs: a semantic annotation file containing object labels and 3D bounding boxes, and a 3D Gaussian Splatting representation. For each scene, we specify the number of target sets, $N_s$, and the number of targets per set, $N_o$. Targets are then selected sequentially until each set contains $N_o$ valid targets.

Annotated objects serve as candidate targets. A candidate is rejected if the current reference position is inside its 3D bounding box or if it is occluded. Visibility is checked by casting a ray toward the object and detecting occupied points within a cylindrical region along the ray. For each valid candidate, a navigation target point is generated near the object and retained only if its distance from the current position lies between $2\mathrm{m}$ and $10\mathrm{m}$. Each accepted target point becomes the reference position for the next selection.

After all $N_s$ target sets are generated, the target sequences are passed to GCOPTER~\cite{WANG2022GCOPTER} to optimize smooth, collision-free trajectories subject to UAV motion constraints. Along each trajectory, RGB images and estimated depth maps are rendered from the 3D Gaussian Splatting representation using GSplat and recorded together with the corresponding UAV poses.

\begin{figure*}[!t]
  \centering
    \includegraphics[width=1\linewidth]{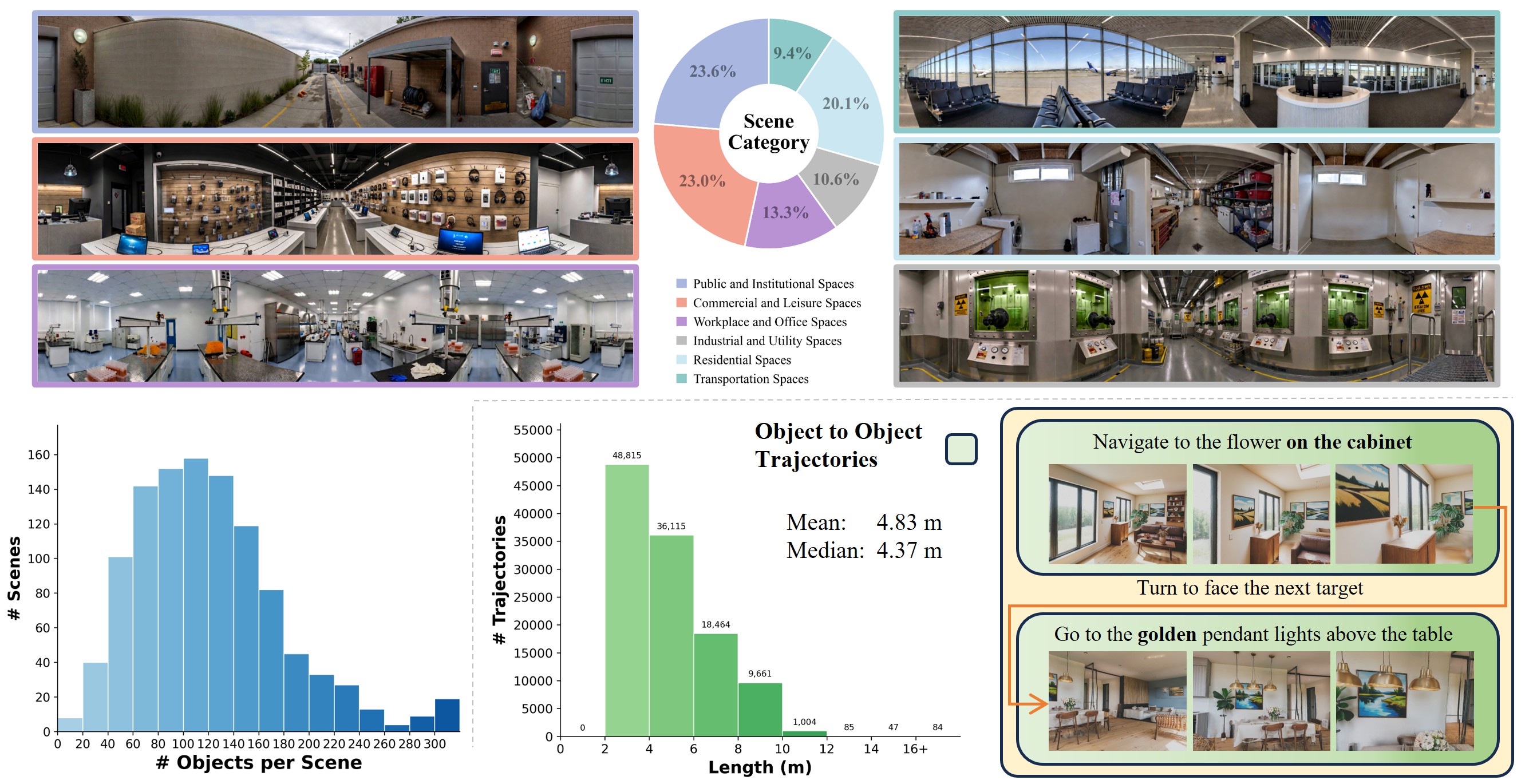}
    \caption{\textbf{Dataset Statistics}. Top is the distribution of scene categories. Bottom left is object count distribution in generated scenes. Bottom right is trajectory statistics \& example trajectory. }
  \label{fig:scene stat}
\end{figure*}

\begin{table*}[htb]
\captionsetup{justification=centering}
\centering
\caption{Trajectory Datasets Statistical Comparison}
\label{tab:comparison}
\resizebox{\textwidth}{!}{
\begin{tabular}{l c c c c c c c c}
\toprule
\textbf{Dataset} 
& $N_{\text{traj}}$ 
& Action Space 
& $N_{\text{scenes}}$ 
& Scene Environment 
& Traj Generation 
& Traj Annotation 
& Kinematics 
& Scene/Traj. Scaling \\
\midrule

R2R\cite{mattersim}  
& 7189 & Node-based & 90 & Matterport3D 
& Sample from Nodes & Human & No 
& Manual/\textbf{Auto} \\

RxR\cite{rxr}  
& 13992 & Node-based & 90 & Matterport3D 
& Sample from Nodes & Human & No 
& Manual/\textbf{Auto} \\

VLN-CE\cite{krantz_vlnce_2020} 
& 4475 & 2 DoF & 90 & Matterport3D 
& A* between Nodes & from R2R & No 
& Manual/\textbf{Auto} \\

LHPR-VLN\cite{song2024towards} 
& 3260 & 2 DoF & 216 & HM3D 
& D* Lite-based & \textbf{LLM-based} & No 
& Manual/\textbf{Auto} \\

SAGE-3D\cite{miao2025physicallyexecutable3dgaussian} 
& \textbf{100K+*} & 2 DoF & 1000 & \textbf{3DGS} 
& A* & \textbf{LLM-based} & No 
& Manual/\textbf{Auto} \\

\midrule

AVDN\cite{fan-etal-2023-aerial} 
& 6269 & 3 DoF & - & xView 
& Human Control & Human & No 
& -/Manual \\

AerialVLN\cite{liu_2023_AerialVLN} 
& 8446 & 4 DoF & 25 & AirSim + UE 
& Human Control & Human & No 
& Manual/Manual \\

CityNav\cite{lee2024citynav}  
& 32637 & 4 DoF & 34 & SensatUrban 
& Human Control & Pre-existing & No 
& Manual/Manual \\

OpenUAV\cite{wang2024realisticuavvisionlanguagenavigation} 
& 12149 & \textbf{6 DoF} & 22 & AirSim + UE 
& Human Control & \textbf{LLM-based} & \textbf{Yes} 
& Manual/Manual \\

OpenFly\cite{OpenFly} 
& 100K & 4 DoF & 18 & AirSim/\textbf{3DGS}/etc. 
& A* & \textbf{LLM-based} & No 
& Manual/\textbf{Auto} \\

UAV-Flow\cite{wang2025uavflowcolosseorealworldbenchmark}  
& 40801 & \textbf{6 DoF} & - & Real World/UnrealCV 
& Human Control & Human+LLM & \textbf{Yes} 
& -/Manual \\

IndoorUAV\cite{liu2026indooruav} 
& 34925 & 4 DoF & 1075 & MP3D/HM3D/etc. 
& Human Control & \textbf{LLM-based} & No 
& Manual/Manual \\

\midrule

FlyMirage 
& \textbf{110K+} & \textbf{6 DoF} & \textbf{1100} & \textbf{3DGS} 
& \textbf{Dynamics-Aware Planner} & \textbf{LLM-based} & \textbf{Yes} 
& \textbf{Auto/Auto} \\

\bottomrule
\end{tabular}
}
\caption*{\scriptsize \textbf{*}: Despite SAGE-3D providing 2 million instruction-trajectory pairs, each trajectory is associated with 15--20 prompts, resulting in approximately 100K+ unique trajectories.}

\end{table*}

Within the collected flight data, we sample RGB images from each trajectory to query Qwen-3.5-Flash for automated visual quality assessment, filtering out trajectories with severe rendering artifacts, poor visibility, or unclear flight intent. Only trajectories that are validated as high-quality are archived in FlyMirage. For each valid trajectory, we further query Qwen-3.5-Flash to generate corresponding instructions.

\section{Experiments}

\subsection{Dataset Analysis}

Our dataset contains 1100 3DGS scenes spanning six categories as shown in Fig. \ref{fig:scene stat}: public and institutional spaces; commercial and leisure spaces; workplace and office spaces; industrial and utility spaces; residential spaces; transportation spaces. Existing aerial VLN datasets are often collected within a single environment category. For example, HM3D \cite{ramakrishnan2021hm3d} used by IndoorUAV is primarily home-focused, while approximately 75\% of scenes in SAGE-3D are residential.  In contrast, FlyMirage provides a more balanced and diverse distribution of scene types as shown in Fig. \ref{fig:scene stat}, making it better suited for training models that can generalize across a wider range of environments beyond residential spaces.

Across these scenes, more than 10,000 unique object labels are automatically identified, along with associated bounding boxes for each object instance. A typical scene in this dataset contains 80-120 object instances, providing sufficient semantic context for downstream navigation. Although other datasets, such as InteriorGS, also contain many object instances, a large portion of these instances are repetitive, resulting in only around 700 unique object categories across all scenes of the dataset.

Using this data, we collect over 110,000 navigation trajectories, primarily between object-centric waypoints, achieving the largest scale among the compared datasets, as shown in Table~\ref{tab:comparison}. In addition, existing aerial datasets generally trade off automation and motion feasibility. UAV-Flow provides kinematically feasible trajectories but relies on manual human control for trajectory collection, while OpenFly automates trajectory generation using A* without explicitly considering UAV kinematics. In contrast, FlyMirage uses GCOPTER~\cite{WANG2022GCOPTER} to provide dynamically optimized motion. Most importantly, this is the only aerial dataset in Table~\ref{tab:comparison} that fully automates the acquisition of both additional scenes and trajectories, without requiring manual effort.

The mean and median trajectory lengths are 4.83 m and 4.37 m. As we also record up to five navigation tasks within a continuous run, trajectories can be potentially composed into long-horizon tasks of up to approximately four to five times the typical path length. Moreover, the monetary cost and human effort required to generate our dataset are extremely low. The total API cost, including Marble and all LLM calls, is approximately \$2 per scene, and batch rendering with GSplat requires only a consumer-grade NVIDIA RTX 4070 GPU. From initialization to trajectory collection, the full pipeline takes about one hour per scene and can be parallelized across batches. In comparison, commonly used simulation-reconstruction pipelines such as HM3D require a dedicated Matterport Pro2 sensor, which costs approximately \$3,000, to scan an environment, while SAGE-3D requires manual scene design for each individual scene. 

To further evaluate the quality of FlyMirage, we focus on two fundamental components of navigation data: trajectories and visual observations. We assess trajectory quality through normalized curvature variation~\cite{s23177421}, which measures trajectory smoothness, and evaluate visual observations using the DINOv2~\cite{oquab2023dinov2} Vendi Score~\cite{friedman2023vendi}, which measures visual diversity.

\begin{figure}[htb]
\centering
\includegraphics[width=0.9\linewidth]{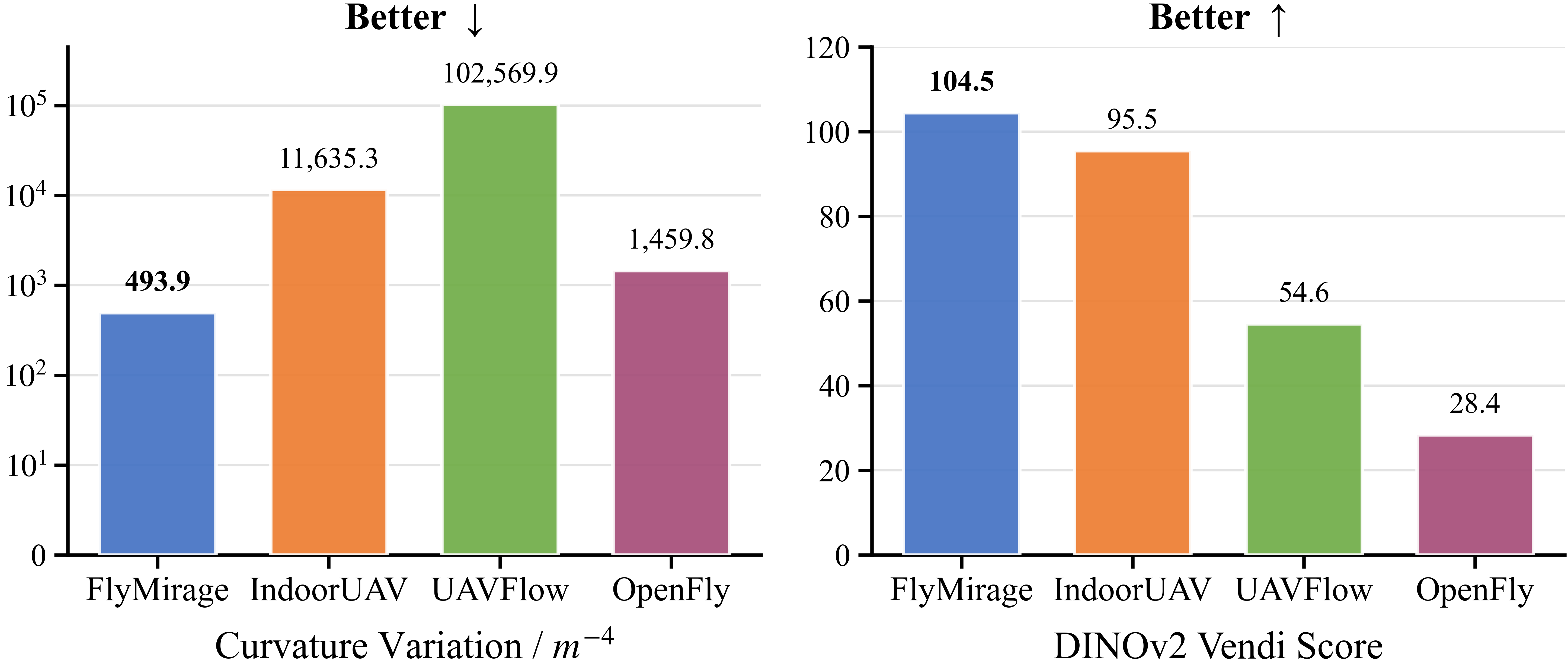}
\caption{\textbf{Dataset Metric Comparison}. Two metrics are used to assess the smoothness and visual diversity.}
\label{fig:dataset_metrics}
\end{figure}

Normalized curvature variation measures how rapidly trajectory curvature changes along the path:
\begin{equation}
E_{\mathrm{cv}} =
\frac{1}{L}\int_{0}^{L}
\left(\frac{d\kappa(s)}{ds}\right)^2 ds,
\end{equation}
where $\kappa(s)$ denotes the curvature at arc-length position $s$ and $L$ is the total trajectory length. Lower values indicate smoother and more gradually changing trajectories. As shown in Fig.~\ref{fig:dataset_metrics}, FlyMirage achieves the lowest curvature variation of $493.9$, approximately $23.6\times$ lower than IndoorUAV. IndoorUAV relies on manually collected trajectories using a limited set of discrete control commands, which can introduce local fluctuations in flight direction. In contrast, GCOPTER optimization enables FlyMirage to generate substantially smoother trajectories, providing better demonstrations for the navigation model to learn from.

We evaluate the visual observations using the DINOv2 Vendi Score. Given a set of L2-normalized DINOv2 image embeddings, the score is computed as the exponential spectral entropy of their feature distribution:
\begin{equation}
\mathrm{VS} =
\exp\left(-\sum_i \lambda_i \log \lambda_i\right),
\end{equation}
where $\lambda_i$ denotes a normalized eigenvalue of the embedding covariance matrix. Higher values indicate greater effective visual diversity. As shown in Fig.~\ref{fig:dataset_metrics}, FlyMirage achieves the highest Vendi Score of $104.5$, exceeding IndoorUAV by $9.5\%$ and UAV-Flow by $91.2\%$. This improvement is consistent with the broader scene distribution of FlyMirage: IndoorUAV is largely derived from residential environments, while UAV-Flow's real-world data are primarily collected on university campuses with a more limited range of architectural styles. In contrast, FlyMirage spans six functional scene categories, providing broader visual and semantic variation across its navigation observations.

\subsection{Baseline Model}
As datasets ultimately serve as the foundation for model training, we validate the effectiveness of our dataset by fine-tuning the widely adopted generalist Vision-Language-Action (VLA) model $\pi_{0}$~\cite{black2026pi0visionlanguageactionflowmodel}. The model combines a pretrained PaliGemma vision-language backbone with a dedicated action expert for generating control actions. Given visual observations, language instructions, and the current agent state, it employs conditional flow matching to generate a chunk of future actions.

Formally, at time step $t$, the model receives the natural-language instruction $\mathcal{I}$, the available visual observations $\mathcal{O}_{1:t}$, and the current agent state
$s_t=(x_t,y_t,z_t,\theta_t)$. It predicts a horizon of $h$ future relative displacements:
\begin{equation}
    \Delta\hat{S}_{t+1:t+h}
    =
    \operatorname{Model}
    \left(\mathcal{O}_{1:t},\mathcal{I},s_t\right),
\end{equation}
where
$\Delta\hat{S}_{t+1:t+h}
=\{\Delta\hat{s}_{t+1},\ldots,\Delta\hat{s}_{t+h}\}$
and each predicted displacement consists of a three-dimensional position offset and a relative yaw angle,
$\Delta\hat{s}_{t+i}=(\Delta\hat{x}_{t+i},\Delta\hat{y}_{t+i},
\Delta\hat{z}_{t+i},\Delta\hat{\theta}_{t+i})$.

\subsection{Simulation Experiment}

\subsubsection{Evaluation Metrics}
\paragraph{Success Rate (SR) \& Oracle Success Rate (OSR)}
A trajectory is successful if its final position is within $0.5\mathrm{m}$ of the target and its yaw error is below $\pi/4$. OSR counts a trajectory as successful if any point along it satisfies this criterion.

\paragraph{Normalized Bending Energy (NBE)}
Normalized bending energy measures the geometric smoothness of a trajectory based on its curvature. Given a trajectory with arc length $s$ and total length $L$, it is defined as
\begin{equation}
E_{\mathrm{bend}}
=
\frac{1}{L}
\int_{0}^{L}
\kappa(s)^2 ds,
\end{equation}
where $\kappa(s)$ denotes the three-dimensional curvature of the trajectory. A lower normalized bending energy indicates a smoother trajectory with less pronounced bending and fewer tight turns.

\subsubsection{Evaluation Dataset}
To evaluate cross-domain generalization and mitigate dataset-specific evaluation bias, we construct an evaluation split of around 200 trajectories in ScanNet++~\cite{yeshwanthliu2023scannetpp} environments. ScanNet++ provides high-fidelity real-world scenes captured using sub-millimeter-resolution laser scans and high-resolution DSLR images. All evaluation trajectories are collected by human pilots, with the corresponding language instructions manually annotated.

\begin{table}[htb]
\centering
\caption{Simulation and real-world evaluation}
\label{tab:dataset_full_real}
\small
\setlength{\tabcolsep}{4pt}
\renewcommand{\arraystretch}{1.08}

\resizebox{\linewidth}{!}{
\begin{tabular}{l*{5}{c}}
\toprule
& \multicolumn{3}{c}{ScanNet++ (Simulation)}
& \multicolumn{1}{c}{Indoor (Real)}
& \multicolumn{1}{c}{Outdoor (Real)} \\
\cmidrule(lr){2-4}
\cmidrule(lr){5-5}
\cmidrule(lr){6-6}

Dataset
& SR/\%$\uparrow$
& OSR/\%$\uparrow$
& NBE/$m^{-2}$$\downarrow$
& SR/\%$\uparrow$
& SR/\%$\uparrow$ \\
\midrule

OpenFly
& 0.0 & 2.1 & N/A$^{*}$
& N/A$^{*}$ & N/A$^{*}$\\

UAV-Flow
& 1.5 & 5.1 & N/A$^{*}$
& N/A$^{*}$ & N/A$^{*}$ \\

IndoorUAV
& 23.6 & 29.2 & 3.7
& 30.0 (3/10) & 10.0 (1/10) \\

FlyMirage
& \textbf{55.4} & \textbf{75.9} & \textbf{1.6}
& \textbf{50.0 (5/10)} & \textbf{50.0 (5/10)} \\

\bottomrule
\end{tabular}
}

\caption*{\scriptsize \textbf{*}: Due to the low success rate, NBE does not yield a reliable value. For the same reason, real-world experiments are not conducted due to safety concerns.}
\end{table}
\vspace{-1em}

As shown in Table~\ref{tab:dataset_full_real}, we compare models trained on four different datasets. The FlyMirage-trained model substantially outperforms those trained on the other datasets on both the ScanNet++ evaluation and real-world experiments, demonstrating stronger cross-domain generalization. Given the limited availability of aerial VLN datasets, OpenFly and UAV-Flow are among the closest publicly available datasets to our setting, although their original task formulations are not fully aligned with ours. In particular, UAV-Flow emphasizes fine-grained language control rather than general navigation across diverse environments, which may limit its transferability under our unified training and evaluation setting. Beyond navigation success, the FlyMirage-trained model also achieves a lower NBE, indicating smoother paths towards the target with fewer sharp bends and abrupt directional changes. Such trajectories are potentially easier for a physical UAV to track during real-world deployment.

\subsection{Real World Experiment}
For the real-world experiments, we run $\pi_0$ inference on a server equipped with an NVIDIA H20 GPU. At each inference step, the quadcopter transmits its current visual observation and odometry information to the server, where $\pi_0$ predicts the next waypoint. This waypoint is then sent back to the quadcopter and executed using EGO-Planner~\cite{9309347}, which performs real-time collision-aware trajectory planning. As the quadcopter approaches the received waypoint, it captures a new observation and sends it to the server to trigger the next inference step. This closed-loop procedure is repeated until the navigation task is completed.

\begin{figure}[htb]
  \centering
    \includegraphics[width=1.0\linewidth]{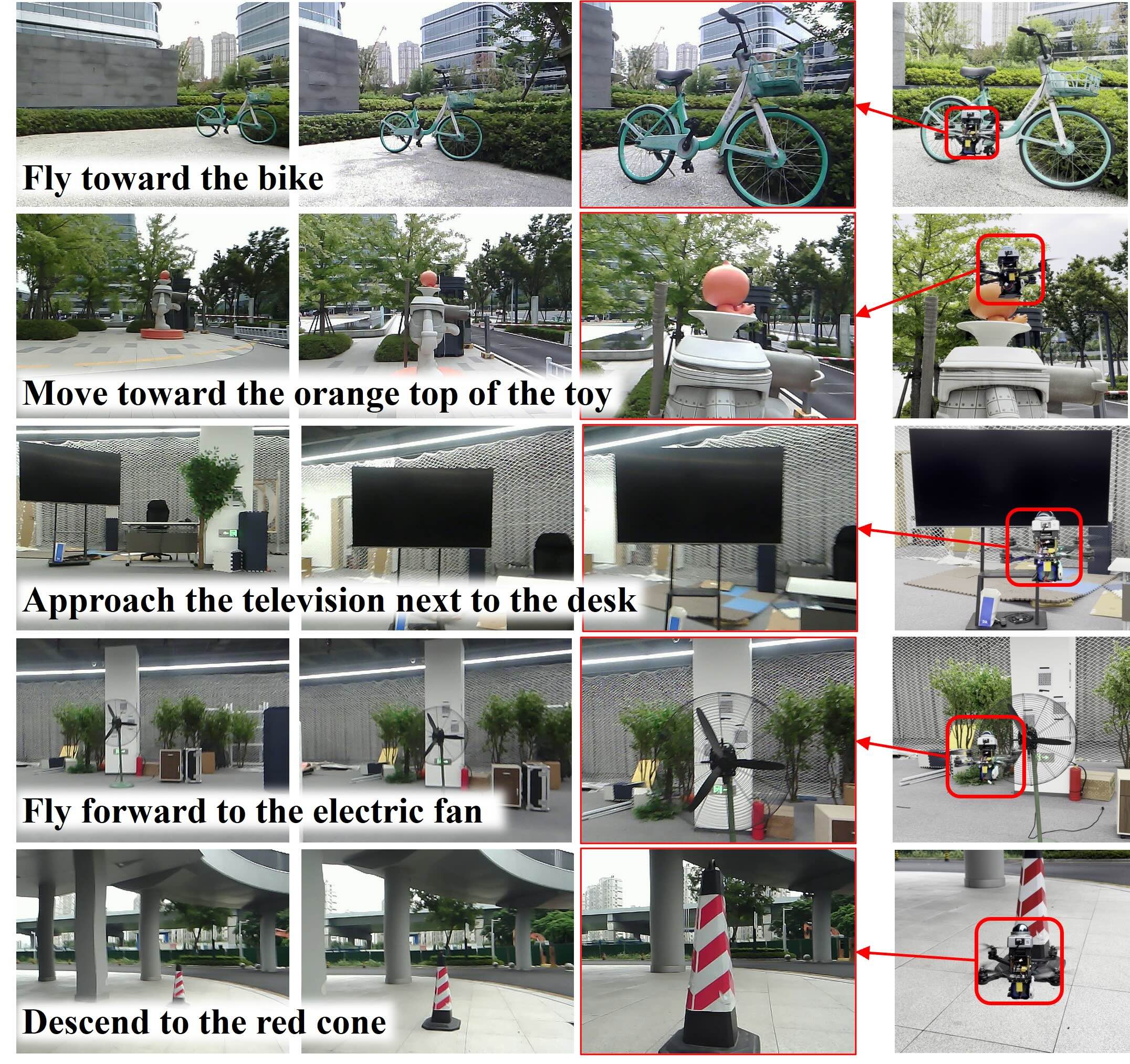}
    \caption{\textbf{Real-World Experiments}. Five examples of real-world experiments performed by FlyMirage-trained model.}
  \label{fig:real_world}
  \vspace{-1em}
\end{figure}

We deploy the IndoorUAV- and FlyMirage-trained models, which achieve sufficiently high simulation success rates for safe real-world evaluation. As shown in Table~\ref{tab:dataset_full_real}, the FlyMirage-trained model achieves a higher success rate in indoor environments, completing 5/10 trials compared with 3/10 for the IndoorUAV-trained model. This improvement suggests that the greater scene and trajectory diversity of FlyMirage enables the model to handle a wider range of navigation targets and language instructions, even in indoor settings. More importantly, FlyMirage compensates for the limited environmental coverage of IndoorUAV by extending the learned navigation capability to outdoor environments. The FlyMirage-trained model completes 5/10 outdoor trials, whereas the IndoorUAV-trained model succeeds in only 1/10. These results demonstrate that the diversity of FlyMirage improves both navigation across different commands and generalization beyond indoor environments.

\subsection{Ablation Studies}

\begin{table}[htb]
\centering
\caption{Ablation study on dataset scale with all datasets matched to $\sim$35K trajectories.}
\label{tab:dataset_matched}
\small
\setlength{\tabcolsep}{8pt}
\renewcommand{\arraystretch}{1.08}

\begin{tabular}{lcc}
\toprule
Dataset
& SR \% $\uparrow$
& OSR \% $\uparrow$ \\
\midrule
OpenFly              & 0.0          & 2.1   \\
UAV-Flow             & 1.0          & 4.1   \\
IndoorUAV            & 23.6         & 29.2  \\
FlyMirage   & \textbf{48.7} & \textbf{69.2} \\
\bottomrule
\vspace{-1em}
\end{tabular}
\end{table}

We conduct ablation studies on dataset scale and diversity. For scale, we match all four training datasets to approximately 35K trajectories and train under identical settings. FlyMirage still outperforms the other datasets, demonstrating the advantage of its high-quality observations and dynamically feasible trajectories. However, it underperforms the full-scale FlyMirage model, showing that dataset scale also contributes to the gains. For diversity, we restrict FlyMirage to residential scenes. As shown in Table~\ref{tab:quality_results}, this reduces the success rate by 10.3 percentage points, highlighting the importance of scene diversity for cross-environment generalization.

\begin{table}[htb]
\centering
\caption{Ablation study on dataset diversity}
\label{tab:quality_results}
\small
\setlength{\tabcolsep}{8pt}
\renewcommand{\arraystretch}{1.08}

\begin{tabular}{lcc}
\toprule
Dataset
& SR \% $\uparrow$
& OSR \% $\uparrow$ \\
\midrule
Residential Scenes Only & 45.1 & 69.2 \\
Original Full Dataset     & \textbf{55.4} & \textbf{75.9} \\
\bottomrule
\end{tabular}
\vspace{-1em}

\end{table}

\section{Conclusion \& Future Work}
In this paper, we present FlyMirage, an automated pipeline that bridges scale and realism in aerial Vision-Language Navigation. Using it, we generate a dataset with 1100 unique scenes and over 110K dynamically feasible UAV trajectories, supporting aerial VLA/VLN~\cite{wu2025vlaanefficientonboardvisionlanguageaction}. In the future, this scalable toolchain and dataset could potentially provide a strong foundation for training next-generation embodied agents or action world models for aerial navigation~\cite{navdreamer}.

\bibliographystyle{IEEEtranBST/IEEEtran}
\bibliography{IEEEtranBST/IEEEabrv,references}

\end{document}